\documentclass[twoside,journal]{IEEEtran}
\usepackage{amsmath,amsfonts}
\usepackage{array}
\usepackage{textcomp}
\usepackage{stfloats}
\usepackage{url}
\usepackage{verbatim}
\usepackage{graphicx}
\usepackage{cite}
\usepackage{subcaption}
\usepackage{booktabs}
\usepackage{rotating}
\usepackage{multirow}
\usepackage{url}
\usepackage[hidelinks]{hyperref}
\usepackage{multirow}
\usepackage[table]{xcolor}
\usepackage{amsmath,amssymb}
\usepackage[ruled,linesnumbered]{algorithm2e} % 不要再加载 algorithm/algpseudocode
\SetKwInOut{Input}{Input}
\SetKwInOut{Initialize}{Initialize}

\begin{document}
\title{What Makes You Unique? Attribute Prompt Composition for Object Re-Identification}
%\title{Attribute Prompt Composition for Object Re-Identification}
\author{Yingquan Wang, Pingping Zhang, Chong Sun, Dong Wang, Huchuan Lu
\thanks{
Yingquan~Wang, Dong~Wang and Huchuan~Lu are with the School of Information and Communication Engineering, Dalian University of Technology. (Email: yingquan\_w95@mail.dlut.edu.cn; wdice@dlut.edu.cn; lhchuan@dlut.edu.cn)

Pingping~Zhang is with the School of Future Technology, School of Artificial Intelligence, Dalian University of Technology. (Email: zhpp@dlut.edu.cn)

Chong Sun is with the Technical Architecture Department, WeChat AI, Tencent, China. (Email: wangziaidao@gmail.com)}
}

\markboth{IEEE Transactions on Circuits and Systems for Video Technology}{}
\maketitle
\begin{abstract}
Object Re-IDentification (ReID) aims to recognize individuals across non-overlapping camera views.
While recent advances have achieved remarkable progress, most existing models are constrained to either single-domain or cross-domain scenarios, limiting their real-world applicability.
Single-domain models tend to overfit to domain-specific features, whereas cross-domain models often rely on diverse normalization strategies that may inadvertently suppress identity-specific discriminative cues.
To address these limitations, we propose an Attribute Prompt Composition (APC) framework, which exploits textual semantics to jointly enhance discrimination and generalization.
Specifically, we design an Attribute Prompt Generator (APG) consisting of a Semantic Attribute Dictionary (SAD) and a Prompt Composition Module (PCM).
SAD is an over-complete attribute dictionary to provide rich semantic descriptions, while PCM adaptively composes relevant attributes from SAD to generate discriminative attribute-aware features.
In addition, motivated by the strong generalization ability of Vision-Language Models (VLM), we propose a Fast–Slow Training Strategy (FSTS) to balance ReID-specific discrimination and generalizable representation learning.
Specifically, FSTS adopts a Fast Update Stream (FUS) to rapidly acquire ReID-specific discriminative knowledge and a Slow Update Stream (SUS) to retain the generalizable knowledge inherited from the pre-trained VLM.
Through a mutual interaction, the framework effectively focuses on ReID-relevant features while mitigating overfitting.
Extensive experiments on both conventional and Domain Generalized (DG) ReID datasets demonstrate that our framework surpasses state-of-the-art methods, exhibiting superior performances in terms of both discrimination and generalization.
The source code is available at \url{https://github.com/AWangYQ/APC}.
\end{abstract}
%-------------------------------------------------------------------------
\section{Introduction}
\IEEEPARstart{O}{bject} Re-Identification (ReID) is a critical task in computer vision, aiming to identify individuals across multiple non-overlapping camera views.
This task has gained considerable attention due to its potential applications in various real-world scenarios, such as surveillance security, and human behavior analysis.
\begin{figure}[htbp]
    \centering
    \includegraphics[width=0.9\linewidth]{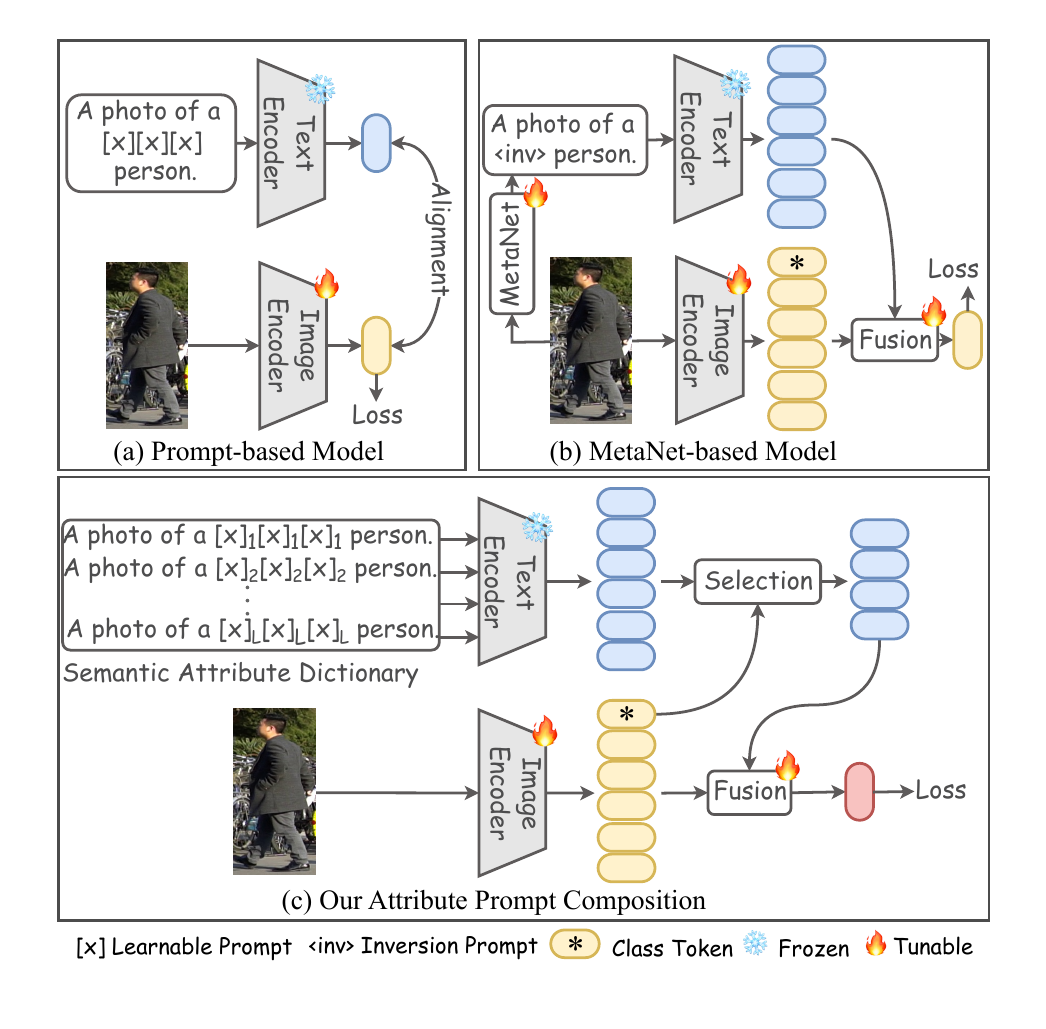}
    \caption{Illustration of various models that employ textual information to obtain discriminative visual representations.}
    \label{fig:1}
\end{figure}

Currently, object ReID methods can be divided into two main categories: single-domain methods and cross-domain methods.
Single-domain methods~\cite{He_2021_ICCV} focus on extracting fine-grained cues from single-domain and have achieved remarkable performances.
While these methods excel in the source domain, they often deliver significant performance degradations when applied to different unseen domains due to the domain shift problem~\cite{Ni_2023_ICCV}.
On the other hand, cross-domain ReID tasks can be divided into Domain Adaptation (DA)~\cite{mekhazni2020unsupervised} and Domain Generalization (DG)~\cite{nie2024latent}.
DA methods can access to target domain data during training, while DG methods require models to generalize to unseen domains without any target domain supervision.
In this paper, we focus on the DG scenario, which better reflects the demands of real-world deployments.
Generally, DG methods utilize advanced techniques such as instance normalization~\cite{nie2024latent}, transfer learning~\cite{wang2024adaptive}, and meta-learning~\cite{ni2022meta} to extract domain-invariant features.
Although these methods have made success in improving generalization, their intricate designs suppress discriminative features, hindering fine-grained identity perception.
Ideally, a robust ReID model should be capable of effective preception in both single-domain and DG scenarios.
However, existing methods typically specialize in only one setting, which limits their applicability in real-world.

In recent years, Contrastive Language-Image Pre-training (CLIP)~\cite{clip} and other Vision-Language Models (VLM) have demonstrated impressive abilities in capturing diverse visual and textual semantic concepts~\cite{vidit2023clip}.
Compared with visual features, language offers a more generalized representation that conveys object distinctions with a reduced sensitivity to domain-specific factors.
Inspired by this, researchers have begun leveraging textual information as a more robust and discriminative form of supervision, enabling models to better generalize across domains.
As a result, numerous CLIP-based ReID models~\cite{li2023clip,yu2024tf,yang2024pedestrian} have been proposed in recent years.
These models are generally divided into two main categories: Prompt-based models and MetaNet-based models, as illustrated in Fig.\ref{fig:1} (a) and Fig.\ref{fig:1} (b), respectively.
Prompt-based models~\cite{li2023clip,yu2024tf} typically learn a set of identity-specific textual prompts, which serve as identity prototypes to supervise the vision encoder.
The impressive results of these models show their strong discrimination and generalization abilities.
However, these methods are primarily reliant on visual representations and do not fully exploit the generalized textual information.
Additionally, learning a distinct prompt for each identity may be redundant and prone to overfitting.
In contrast, MetaNet-based models~\cite{yang2024pedestrian} utilize an inversion network to map image features into the textual space, then feed the inverted image features into a pre-trained text encoder to obtain a generalized textual representation.
Although these models introduce rich textual semantics, the substantial gap between the image and text domains impairs discriminative feature learning and compromises overall performances.

To address the above issues, we propose an Attribute Prompt Composition (APC) framework for object ReID, which includes an Attribute Prompt Generator (APG) and Fast-Slow Training Strategy (FSTS).
As shown in Fig.~\ref{fig:1} (c), the APG models an object as a composition of multiple attributes.
However, there are two key issues in accurately representing different objects through such a conceptual composition: (1) how to construct a comprehensive attribute dictionary relevant to a given object, and (2) how to effectively asses the relevance of each attribute to a specific individual.
To this end, we first introduce a Semantic Attribute Dictionary (SAD), which includes transferable attribute prompts to effectively represent the object’s diverse characteristics.
Unlike existing prompt-learning methods, our method leverages a shared attribute dictionary to encode feature representations, improving generalization across different identities and domains.
Secondly, although multiple objects may be associated with the same attribute prompt, the strength of this association is often instance-specific.
For example, both a 13-year-old person and a 18-year-old person may be described as ``the young'', yet their degrees of association with this attribute differ.
Simply merging attribute prompts without considering their contextual relevance may lead to the loss of fine-grained discriminative cues.
To address this issue, we introduce the Prompt Composition Module (PCM), which extracts the attribute prompts and adaptively aggregates them based on their alignments with the visual representations. This allows the framework to generate more discriminative textual representations.

On the other hand, due to the lack of descriptive annotations in ReID tasks, supervising the visual encoder with one-hot encoded identity labels often leads to catastrophic forgetting.
To address this issue, we propose the FSTS.
Following common ReID practices~\cite{He_2021_ICCV,nie2024latent}, we adopt a Fast Update Stream (FUS) to quickly learn ReID-specific features.
However, this rapid adaptation tends to cause overfitting~\cite{AAformer}.
To mitigate this issue, we introduce a Slow Update Stream (SUS), which employs an exponential moving average strategy to gradually incorporate ReID-specific knowledge while preserving the visual perception ability learned during pre-training.
To transfer the visual perception ability to the FUS, the SUS is employed to construct both visual and attribute-aware prototypes.
These prototypes encode visual information and are used to guide the training of the FUS.
Specifically, we randomly sample one image per identity and feed it into the SUS to extract its visual and attribute-aware features.
These features are used as identity-specific prototypes to supervise the FUS by the contrastive loss.
This collaborative learning approach equips the model with discriminative abilities without compromising the visual perception ability provided by the VLM.
Extensive experiments validate that our proposed framework consistently achieves superior performance and generalization across diverse ReID benchmarks.
% --------------------------------------------------------------------------------

The main contributions are outlined as follows:
\begin{itemize}
\item We propose a novel framework named Attribute Prompt Composition (APC) for object ReID, which uses learnable attribute prompts to generate both discriminative and generalizable representations.
\item We propose an Attribute Prompt Generator (APG), that can represent an object as a composition of multiple attributes and adaptively aggregates relevant attributes.
\item We propose a Fast–Slow Training Strategy (FSTS), that helps learning of ReID-specific knowledge and preserving the visual perception ability from the pre-trained VLM.
\item Extensive experiments verify that the proposed framework achieves state-of-the-art results on both conventional and DG ReID datasets.
\end{itemize}
%-------------------------------------------------------------------------
\section{Related Work}
\subsection{Image-based Object Re-Identification}
As an important application in real-world scenarios, object ReID has garnered long-term attention from both academia and industry.
Early works in ReID primarily employ Convolutional Neural Networks (CNNs) to extract feature representations, and achieve notable success.
For example, Sun~\emph{et al.}~\cite{sun2018beyond} employ horizontal partitioning to learn discriminative local features.
Liu~\emph{et al.}~\cite{liu2020iterative} alternately optimize local fine-grained features and global holistic features to enhance discriminative representations.
Zhang~\emph{et al.}~\cite{alignedreid} not only divide feature maps for learning the local cues but also design a dynamic alignment mechanism to measure the similarity between same semantic parts.
However, CNNs are limited by their tendency to overlook global information, which may lead to overfitting and restrict model performances.
To address the limitations of CNNs, recent studies have incorporated attention mechanisms to enhance feature extraction and improve ReID performances by focusing on relevant parts.
For instance, Xu~\emph{et al.}~\cite{xu2018attention} introduce pose estimation to guide attention generation.
Zhang~\emph{et al.}~\cite{zhang2020relation} generate attention maps in consideration of the pixel relation.
Furthermore, some works have attempted to incorporate attribute priors into the training process.
For example, Zhang~\emph{et al.}~\cite{zhang2023attribute} propose an attribute-guided collaborative learning framework to enhance the robustness of ReID models.
Although the aforementioned works have achieved promising results, they often fail to ensure good performances in the more challenging DG setting.
To this end, Pan~\emph{et al.}~\cite{pan2018two} propose to investigate the impact of combining instance and batch normalization, which has been widely adopted in subsequent DG ReID methods.
However, such normalization techniques inevitably filter out some discriminative information, leading to suboptimal performance in source-domain evaluations.
In contrast, we leverage prompt tuning for domain-robust textual representation learning and introduce a FSTS to balance task adaptation and generalization.
\subsection{Vision-language Learning for ReID}
Recently, large VLM (\emph{e.g.}, CLIP~\cite{clip}) have emerged as a new paradigm for foundational models.
They aim to connect visual representations with their corresponding language descriptions.
To leverage the powerful generalization abilities of the pre-trained VLM for downstream tasks, many works~\cite{menon2023visual,xie2023ra} generate text descriptions based on class names to further supervise model learning.
However, due to the lack of descriptive annotations, adapting the VLM to ReID is not straightforward.
To address this issue, some researchers~\cite{zhai2024multi} utilize off-line Visual Question Answering (VQA) models to generate the captions of each image, enabling image-language supervision.
Although promising results are achieved, the introduced computational overhead is substantial and impractical for real-world applications.
Inspired by the prompt tuning technologies~\cite{zhou2022coop}, some works~\cite{li2023clip,yu2024tf} learn identity-specific prompts for each identity to replace textual descriptions.
For instance, Li \emph{et al.}~\cite{li2023clip} design a two-stage strategy. First, they learn identity-specific prompts, and then they use the prompts to construct a classifier for semantic information transfer.
However, these methods neglect that the rich textual semantic information can complement the visual information for robust representations.
Meanwhile, some works introduce hybrid prompts to enhance the visual representations.
For example, Ma~\emph{et al.}~\cite{ma2023hybridprompt} integrate language model–generated prompts with human priors in prompt tuning, enabling more robust visual question answering.
Similarly, Chen~\emph{et al.}~\cite{chen2024domain} propose a domain-aware multi-modal dialog system that models user characteristics as probability distributions, resulting in more robust cross-domain conversations.
Although these approaches achieve impressive results, they typically learn a separate prompt for each class, which can easily cause overfitting.
Additionally, they learn identity-wise prompts based on a frozen CLIP model, which limits their ability to capture fine-grained discriminative cues.
Different from these methods, we propose an end-to-end framework, that constructs a diverse set of attribute prompts related to the object.
Then, the model selects instance-relevant attributes and adaptively aggregates them into an attribute-aware feature, which effectively preserves discriminative information while reducing domain sensitivity.
%--------------------------------------------------------------------------------------------------------------------------
\section{Proposed Method}
\begin{figure*}[htbp]
\centering
\includegraphics[width=0.9\linewidth]{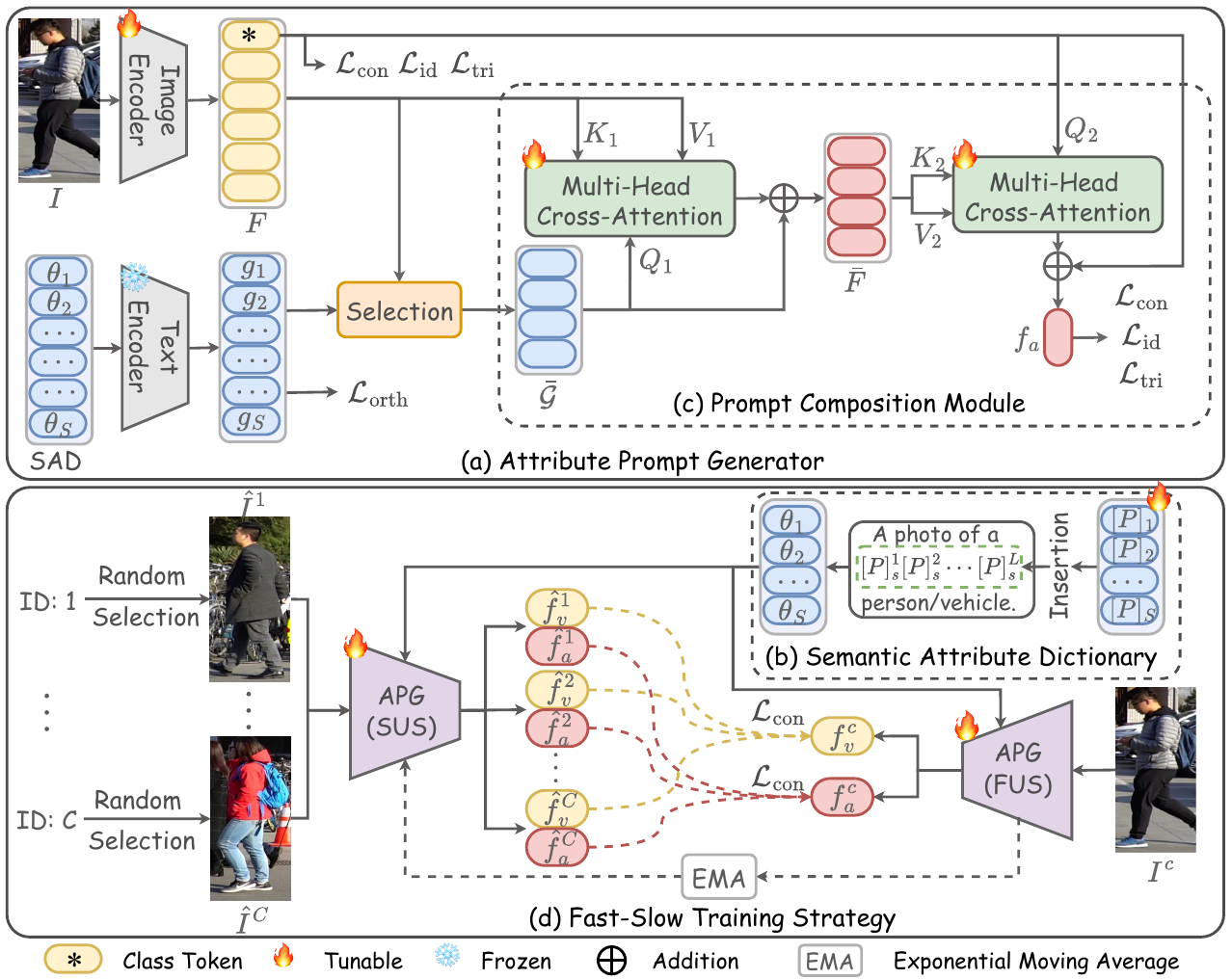}
\caption{
Illustration of the proposed APC framework, comprising APG and FSTS.
APG constructs attribute-aware features by SAD and PCM.
FSTS leverages two streams with different update mechanisms: FUS for ReID-specific knowledge, and SUS for preserving the visual perception ability inherited from VLM.
}
\label{fig:2}
\end{figure*}
\subsection{Overview}
As depicted in Fig.~\ref{fig:2}, we propose an Attribute Prompt Composition (APC) framework, which consists of an Attribute Prompt Generator (APG) and a Fast–Slow Training Strategy (FSTS).
The APG learns comprehensive object-related attributes and composes them into robust attribute-aware representations.
Meanwhile, the FSTS comprises two collaborative streams: the Fast Update Stream (FUS) rapidly captures ReID-specific features, while the Slow Update Stream (SUS) updates more conservatively to retain the visual perception ability from the pre-trained VLM.
The SUS additionally constructs visual and attribute prototypes to guide the FUS.
Through the interaction between FUS and SUS, the framework effectively balances ReID-specific learning and generalization for more robust and discriminative representations.

Technically, we adopt the frozen CLIP text encoder to obtain textual features.
For visual encoding, we use the ViT-B/16 backbone pre-trained by CLIP~\cite{clip} with the side information embedding~\cite{He_2021_ICCV} as the visual encoder $\mathcal{I}(\cdot)$.
Given an image $I \in \mathbb{R}^{H\times W\times D}$, we feed it into the visual encoder to extract visual features:
\begin{equation}
\label{eq.3}
F = \mathcal{I}(I),
\end{equation}
where $H$, $W$ and $D$ denote its height, width and channels respectively.
$F \in \mathbb{R}^{(N+1)\times768}$ is the visual token embedding sequence.
Following the setting of CLIP~\cite{clip}, we project the visual features into a cross-modal embedding space:
\begin{equation}
    R = W_p\cdot F
\end{equation}
where $R$ represents the projected features and $W_p$ is the learnable parameters.
Notably, we utilize the class token of $F$ and $R$ as the visual representation $f_v$ and projected visual representation $r$, respectively.
\subsection{Attribute Prompt Generator}
To introduce VLM to fine-grained perception tasks, some works~\cite{zhou2022coop,li2023clip} utilize identity-specific prompts to guide the vision encoder.
Although these methods have yielded promising results, they tend to overfit when learning a description for each identity.
To address these limitations, we propose the Attribute Prompt Generator (APG), as shown in Fig.~\ref{fig:2} (a).
It integrates SAD and PCM to adaptively compose attribute prompts and obtain robust feature representations.

\textbf{Semantic Attribute Dictionary (SAD).}
Intuitively, an object can be described as a combination of different attributes~\cite{chai2022video}.
Many of these attributes are shared across different objects.
Therefore, we focus on learning discriminative attribute prompts rather than identity-specific prompts.
As shown in Fig.~\ref{fig:2} (b), we construct the SAD and represent each person or vehicle as a combination of attributes.
Specifically, we first insert the learnable tokens into the template as:
\begin{equation}
    \theta_s = \text{``A photo of a }   [\mathrm{P}]_i^{\mathrm{1}}[\mathrm{P}]_i^{\mathrm{2}}
    \cdots[\mathrm{P}]_i^{\mathrm{L}} \text{ person/vehicle.''},
    \label{eq:7}
\end{equation}
where $[\mathrm{P}]^s_{l}$ ($l\in \{1, ..., L\}$, $s\in \{1, ..., S\}$) represents the $l$-\emph{th} learnable token of the $s$-\emph{th} attribute and $S$ denotes the total number of attributes.
Then the SAD is represented as:
\begin{equation}
    \mathcal{H} = \{\theta_1, \theta_2, \cdots, \theta_S\}.
\end{equation}
Afterwards, each attribute prompt is obtained by feeding the attribute into the text encoder $\mathcal{T}(\cdot)$:
\begin{equation}
    g_i = \mathcal{T}(\theta_i),
\end{equation}
Here, $g_i$ refers to the $i$-\emph{th} attribute prompt.
To encourage the diversity among attributes, we introduce an orthogonality loss:
\begin{equation}
    L_{\text{orth}} = ||O-E||_1,
\end{equation}
where $O$ is a similarity matrix with each element $o_{ij} = s(g_i,g_j)$, and $s(\cdot, \cdot)$ denotes the cosine similarity score between attribute prompts $g_i$ and $g_j$. $E$ is the identity matrix, and $||\cdot||_1$ denotes the matrix 1-norm.

Generally, not all attributes from SAD are relevant to the object.
Therefore, we select the relevant attribute and filter out the irrelevant ones.
Specifically, we compute the cosine similarity between the projected visual representation and each attribute prompt, and then select the Top-$K$ most relevant ones,
\begin{equation}
    \bar{\mathcal{G}} = \Bigg\{g_j | j \in Top_K\Big(s(r, g_j)\Big)\Bigg\},
\end{equation}
where $\bar{\mathcal{G}}$ represents the set of selected attribute prompts.
This set is then used for attribute prompt aggregation.

\textbf{Prompt Composition Module (PCM).}
In practice, different objects may have similar attributes, but the relevance of each attribute varies.
Simply concatenating attribute prompts may fail to capture these fine-grained semantic differences.
To address this, we propose PCM, which adaptively aggregates attribute prompts by taking into account their relevance to the visual features.
Specifically, as shown in Fig.~\ref{fig:2} (c), the selected attribute prompts $\bar{\mathcal{G}}$ are passed into a linear transformation to generate a query matrix $Q_1$.
The visual tokens $F$ are passed into two linear transformations to generate a key matrix $K_1$ and a value matrix $V_1$, respectively.
The interaction between the attribute prompts and the visual tokens is represented as:
\begin{equation}
    \bar{F} = \bar{\mathcal{G}} + \sigma(\frac{Q_1K_1^T}{\sqrt{d}})V_1,
\end{equation}
where $\sigma$ is the Softmax function, $(\cdot)^T$ means the matrix transposition, and $d$ denotes the dimension of $Q_1$.
To obtain the final attribute-aware feature, we utilize the projected visual representation $r$ to guide the composition of attribute prompts.
More specifically, $r$ is projected to a query $Q_2$ and $\bar{F}$ is projected to a key matrix $K_2$ and a value matrix $V_2$. The attribute-aware feature can be obtained:
\begin{equation}
    f_a = r + \sigma(\frac{Q_2K_2^T}{\sqrt{d}})V_2.
\end{equation}

Distinct from most prompt-tuning methods~\cite{zhou2022coop}, APG generates a set of shared attribute prompts and adaptively combines them to form robust attribute-aware features.
Moreover, unlike CLIP-ReID~\cite{li2023clip} which encodes objects only using visual features, APG introduces a novel paradigm for object representation by integrating both visual and textual features.
% -----------------------------------------------
\subsection{Fast-Slow Training Strategy}
\label{sec:III.B}
The visual encoders of pre-trained VLM exhibit strong abilities in perceiving comprehensive semantic concepts from images.
A key challenge in leveraging VLM for ReID is that fine-tuning often erodes their general semantic perception ability.
To address this issue, we propose FSTS by duplicating APG into two parallel streams: FUS, denoted as $\mathcal{M}(\cdot)$ and SUS, denoted as $\hat{\mathcal{M}}(\cdot)$.
FUS is employed to capture ReID-specific discriminative features.
Meanwhile, the SUS retains the visual perception abilities inherited from the pre-trained VLM and constructs identity-aware prototypes to transfer this ability to the FUS.
Through joint training, FUS benefits from the semantic guidance provided by SUS, thus achieving better generalization while avoiding overfitting.

To this end, we utilize a exponential moving average strategy~\cite{DINO} to update the parameters of SUS after every few iterations.
The slow update allows SUS to preserve the general semantic perception ability while gradually acquiring ReID-specific knowledge from FUS.
Meanwhile, FUS is updated to rapidly capture ReID-specific cues and guide by general visual perception ability from SUS. The parameter update rules for FUS and SUS are formulated as follows:
\begin{equation}
     \alpha_{\mathcal{M}} \leftarrow \alpha_{\mathcal{M}} - \eta \nabla_{\theta_{\mathcal{M}}} \mathcal{L},
\end{equation}
\begin{equation}
    \alpha_{\hat{\mathcal{M}}} \leftarrow m \cdot \alpha_{\hat{\mathcal{M}}} + (1 - m) \cdot \alpha_{\mathcal{M}},
\end{equation}
where $\alpha_{\mathcal{M}}$ and $\alpha_{\hat{\mathcal{M}}}$ denote the parameters of FUS and SUS, respectively.
$\eta$ is the learning rate, $\mathcal{L}$ is the loss function, and $m$ is the momentum coefficient.
Then, we utilize SUS to construct identity prototypes to store the general visual perception ability inherited from VLM and transfer this ability to FUS by the contrastive loss~\cite{constrastive_loss}.
Specifically, we randomly sample one image $\hat{I}^c$ per identity from the training set.
The selected images are then fed into $\hat{\mathcal{M}}(\cdot)$ to obtain the visual features $\hat{f}^c_v \in \mathbb{R}^{768}$ and the attribute-aware features $\hat{f}^c_a \in \mathbb{R}^{512}$, where $c \in \{1, \cdots, C\}$ and $C$ denotes the total number of identities.
We then construct identity prototypes as follows:
\begin{equation}
    \mathcal{H}_v = [\hat{f}^1_v, \hat{f}^2_v, \cdots, \hat{f}^C_v],
\end{equation}
\begin{equation}
    \mathcal{H}_a = [\hat{f}^1_a, \hat{f}^2_a, \cdots, \hat{f}^C_a].
\end{equation}
As shown in Fig.~\ref{fig:2} (d), when training FUS, for each randomly sampled image $I^c$ belong to the identity $c$, we feed it into $\mathcal{M}(\cdot)$ for visual features and attribute-aware features.
Then, the contrastive loss is used to guide FUS retain generalizable knowledge inherited from VLM:
\begin{equation}
    \begin{split}
                \mathcal{L}_\text{con} = -log \frac{exp(s(\hat{f}^c_v, f^c_v)/\tau)}{\sum^{C}_{j=1} exp(s(\hat{f}^j_v, f^c_v)/\tau)}
     \\ -log\frac{exp(s(\hat{f}^c_a, f^c_a)/\tau)}{\sum^{C}_{i=1} exp(s(\hat{f}^i_a, f^c_a)/\tau)},
    \end{split}
    \label{eq:6}
\end{equation}
where $\tau$ is the temperature parameter.
%-----------------------------------------------------
\begin{algorithm}[t]
    \caption{Training Procedure of Our APC}
    \label{alg:apc}
    \Input{Image $I \in \mathbb{R}^{3\times H\times W}$ }
    \Initialize{Total epochs $E$; Total identities $C$; Total images $K$; Batch-size $B$;
    Index $e = 0$; Iteration $\mathcal{U} = K/B$; Update interval $\tau$}

    \For{$c = 1$ \KwTo $C$}{
    Randomly sample one image $\hat{I}^c$ from the image set with identity $c$\;
    Extract features $\hat{f}^c_v$, and $\hat{f}^c_a$ using SUS via Eq.(1) and Eq.(9)\;
    }
    Construct identity prototypes $\mathcal{H}_v$ and $\mathcal{H}_a$ using Eq.~(12) and Eq.~(13)\;

    \While{$e < E$}{
    \For{$u = 1$ \KwTo $\mathcal{U}$}{
        Extract features $f_v$ using Eq.~(1) and Eq.~(2)\;
        Derive attribute prompts using Eq.~(3)-(5)\;
        Compute the loss $\mathcal{L}_{\text{orth}}$ using Eq.~(6)\;
        Select attribute prompts $\bar{\mathcal{G}}$ with Eq.~(7)\;
        Obtain attribute-aware feature $f_a$ using Eq.~(8) and Eq.~(9)\;
        Calculate the loss $\mathcal{L}_{\text{con}}$ using Eq.~(12)-(14)\;
        Update FUS using Eq.~(10)\;
    }
    \If{$u \bmod \tau = 0$}{
        Update SUS using Eq.~(11)\;
    \For{$c = 1$ \KwTo $C$}{
    Randomly sample one image $\hat{I}^c$ from the image set with identity $c$\;
    Extract features $\hat{f}^c_v$, and $\hat{f}^c_a$ using SUS via Eq.(1) and Eq.(9)\;
    }
    Construct identity prototypes $\mathcal{H}_v$ and $\mathcal{H}_a$ using Eq.~(12) and Eq.~(13)\;
    }
    $e \gets e+1$\;
    }
    \Return{} SUS
\end{algorithm}

The proposed FSTS shares certain similarities with memory bank methods~\cite{dai2022cluster,he2020momentum}, but fundamentally differs in its design philosophy.
Previous memory bank methods aim to enlarge the batch size, while FSTS uses a slow update stream to simultaneously obtain ReID-specific knowledge and preserve the visual perception ability inherited from the pre-trained VLM.
This enables the model to improve its discriminative ability while effectively mitigating overfitting.
\subsection{Optimization}
We adopt the following loss function to optimize our model,
\begin{equation}
    \begin{split}
        \mathcal{L} = \mathcal{L}_{\text{id}}(f_a) + \mathcal{L}_{\text{id}}(f_v) + \mathcal{L}_{\text{tri}}(f_a)\\ + \mathcal{L}_{\text{tri}}(f_v) +  \mathcal{L}_\text{con}+ \kappa \mathcal{L}_{\text{orth}},
    \end{split}
\end{equation}
where $\kappa$ is set to $0.1$.
$\mathcal{L}_{id}$ and $\mathcal{L}_{tri}$ are the identity loss~\cite{CEloss} and triplet loss~\cite{tripletloss}, respectively.
The training procedure of our APC is shown in Alg.~\ref{alg:apc}.
During inference, the attribute-aware features $f_a$ and the visual representations $f_v$ are concatenated to form the final object representation.
\section{Experiments}
\subsection{Datasets and Protocols}
\label{section:Datasets and Protocols}
We conduct extensive experiments on several real and synthetic image-based object ReID datasets.
More specifically, MSMT17, Market1501, DukeMTMC, CUHK03-NP and VeRi-776 are sampled in the real, while RandPerson is a large-scale synthetic dataset.
The details are summarized in Tab.~\ref{table:datasets}.
\begin{table}[htbp]
      \centering
      \renewcommand\arraystretch{1.1}
      \setlength\tabcolsep{5.5pt}
      \caption{Statistics of used datasets.}
      \resizebox{0.46\textwidth}{!}{
      \begin{tabular}{cccc}
          \hline
           Dataset              &Identities    &Images  &Cameras(View)\\
           \hline
           MSMT17~\cite{MSMT17}              &4,101  &126,441   &15\\
           Market1501~\cite{market1501}          &1,501  &32,668    &6\\
           DukeMTMC~\cite{Duke}       &1,404  &36,441    &8\\
           CUHK03-NP~\cite{cuhk03}           &1,467    &28,192    &2\\
           VeRi-776~\cite{VeRi}                  &776     &49,357    &20(8)\\
           RandPerson~\cite{RandPerson}          &8,000    & 132,145  &19\\
           \hline
      \end{tabular}}
      \label{table:datasets}
  \end{table}

To validate the representation ability and generalization of our model, we evaluate its performances in both single-domain ReID and DG ReID settings.
Specifically, we train and test the model on the same dataset for the single-domain setting.
In the DG setting, we employ three testing settings to assess the model's generalization ability~\cite{Ni_2022_CVPR}.
The first setting involves training on one dataset and testing on another unseen dataset.
The second setting utilizes the training sets of multiple datasets for training and evaluates on the test set of an unseen dataset.
The third setting trains on a large-scale synthetic dataset and tests on other real-world datasets.
For simplicity and clarity, we denote Market1501, DukeMTMC, CUHK03-NP and MSMT17 as \textbf{M}, \textbf{D}, \textbf{C} and \textbf{MS}, respectively.
Performances are evaluated by mean Average Precision (mAP) and Cumulative Matching Characteristic at Rank-1 (R1).
\subsection{Implementation Details}
The proposed method is implemented using PyTorch.
We adopt CLIP-ViT-B/16 with a patch stride of 12 as the image encoder.
The batch size is set to 64, consisting of 16 identities with 4 images per identity.
The input images from person datasets are resized to $256 \times 128$, while those from the vehicle dataset are resized to $256 \times 256$.
All images are augmented with random cropping, horizontal flipping, and random erasing~\cite{randomerasing}.
The Adam optimizer~\cite{kingma2014adam} is adopted with an initial learning rate of $7.5\times 10^{-6}$, and the model is trained for 60 epochs using a cosine learning rate decay schedule.
For all datasets, the temperature parameter $\tau$ and momentum coefficient are set to 0.007 and 0.0004, respectively.
Each attribute prompt comprises four learnable tokens.
\subsection{Comparison with State-of-the-Arts}
In Tab.~\ref{table:image-based sota}, Tab.~\ref{tabel:vehicle sota} and Tab.~\ref{table:DG sota}, we compare the proposed method with state-of-the-art methods on traditional person ReID, vehicle ReID and DG person ReID.
The experimental results show that our method consistently outperforms most existing methods across all tasks.
It is worth noting that no post-processing techniques are applied to the reported results.

\textbf{Traditional Object ReID.}
To verify the effectiveness of the proposed method, we conduct experiments on four commonly used object ReID benchmarks, as shown in Tab.~\ref{table:image-based sota} and Tab.~\ref{tabel:vehicle sota}.
On Market1501, our method achieves the best performances in mAP and comparable performances in R1.
On MSMT17, our method significantly outperforms previous methods (\emph{e.g.}, CLIP-ReID) with a notable improvement in mAP (+1.3\%).
On DukeMTMC, our method also demonstrates superiority, surpassing CLIP-ReID in both mAP (+1.0\%) and R1 (+1.1\%).
On VeRi-776, our method achieves 97.9\% in R1 and 85.1\% in mAP, outperforming CLIP-ReID by 0.6\% in both metrics.
\begin{table}[htbp]
\renewcommand\arraystretch{1.1}
\setlength\tabcolsep{5.5pt}
    \caption{Quantitative comparison of state-of-the-art methods on three single-domain person Re-ID datasets. The best and second-best results are in \textbf{bold} and \underline{underlined}, respectively.}
    \centering
    \resizebox{0.46\textwidth}{!}{
    \begin{tabular}{l|cccccc}
    \hline
    \multirow{2}{*}{Method}&
    \multicolumn{2}{c}{Market1501}&
    \multicolumn{2}{c}{MSMT17}&
    \multicolumn{2}{c}{DukeMTMC}\\
                &             mAP         &R1         & mAP         & R1         & mAP         & R1\\
    \hline
    Nformer~\cite{wang2022nformer}           &\underline{91.1}        &94.7          &59.8         &77.3           &{\underline{83.5}}         &89.4\\
    AAformer~\cite{AAformer}       &87.7        &95.4          &62.6         &83.1           &80.0         &90.1\\
    TransReID~\cite{He_2021_ICCV}       &88.9        &95.2          &67.4         &85.3           &82.0         &{{90.7}}\\
    APD~\cite{lai2021transformer}            &87.5        &95.5          &57.1         &79.8           &74.2         &87.1\\
    HAT~\cite{zhang2021hat}              &89.8        &{{95.8}}          &61.2         &82.3           &81.4         &90.4\\
    ADSO~\cite{Zhang_2021_CVPR}     &87.7   &94.8   &--  &--  &74.9  &87.4\\
    PFD~\cite{wang2022pose}            &89.6        &95.5          &65.1         &82.7           &{82.2}         &90.6\\
    DCAL~\cite{Zhu_2022_CVPR}            &87.5        &94.7          &64.0         &83.1           &80.1         &89.0\\
    SAP~\cite{jia2023semi}             &{90.5}        &{\underline{96.0}}          &67.8         &85.7           &--     &--\\
    DC-Former~\cite{li2023dc}      &90.4        &{\underline{96.0}}          &{68.8}         &{86.2}           &--     &--\\
    RGANet~\cite{he2023region}  &89.8 &95.5 &{{72.3}} &{{88.1}} &--   &--\\
    PHA~\cite{Zhang_2023_CVPR} &90.2 &{\textbf{96.1}} &68.9 &86.1 &-- &-- \\
    CLIP-ReID~\cite{li2023clip} &{90.5} &95.4 &{\underline{75.8}} &{\underline{89.7}} &{{83.1}} &{\underline{90.8}}\\
    \hline
    Ours          &{\textbf{91.2}}        &{\underline{96.0}}         &{\textbf{77.1}}       &{\textbf{90.1}}         &{\textbf{84.1}}         &{\textbf{91.9}}\\
    \hline
    \end{tabular}}
    \label{table:image-based sota}
\end{table}
%--------------------------------------------------
\begin{table}[htbp]
\renewcommand\arraystretch{1.1}
\setlength\tabcolsep{5.5pt}
      \caption{Quantitative comparison of state-of-the-art methods on VeRi-776.}
    \centering
    \resizebox{0.46\textwidth}{!}{
    \begin{tabular}{l|c|cc}
    \hline
    \multirow{2}{*}{Method}&
    \multirow{2}{*}{Source}&
    \multicolumn{2}{c}{VeRi-776}\\
                 &           &mAP         &Rank1\\
    \hline
    PGAN~\cite{zhang2020part}        &TITS 2020     &79.3        &96.5\\
    PVEN~\cite{meng2020parsing}       &CVPR 2020      &79.5        &95.6\\
    SAVER~\cite{khorramshahi2020devil} &ECCV 2020           &79.6        &96.4\\
    CFVMNet~\cite{sun2020cfvmnet}      &MM 2020    &77.1        &95.3\\
    GLAMOR~\cite{suprem2020looking}     &ArXiv 2020      &80.3        &96.5\\
    MPC~\cite{li2021exploiting}   &TMM 2021  &{80.9}  &96.2\\
    MsKAT~\cite{li2022mskat}    &TITS 2022    &{82.0}  &{{97.1}}\\
    TransReID~\cite{He_2021_ICCV}  &ICCV 2021    &{82.0}       &{{97.1}}\\
    DCAL~\cite{Zhu_2022_CVPR}     &CVPR 2022     &80.2        &{96.9}\\
    SOFCT~\cite{SOFCT}  &TITS 2023 &80.7  &96.6\\
    GSE-Net~\cite{GSE-Net} &TITS 2024  &81.3  &96.3\\
    ARPM-TransReID~\cite{ARPM-TransReID} &ICASSP 2025 &81.4 &97.0\\
    ADPRP-Net~\cite{ADPRP-Net}  &PR 2025    &{{82.8}}  &95.6\\
    CLIP-ReID~\cite{li2023clip}  &AAAI 2023  &{\underline{84.5}}  &{\underline{97.3}}\\
    \hline
    Ours  &   &{\textbf{85.1}}        &{\textbf{97.9}}\\
    \hline
    \end{tabular}}
    \label{tabel:vehicle sota}
\end{table}
%----------------------------------------

\textbf{Single-source DG ReID.}
Single-source is the most common setting in real-world scenarios, which trains on one single dataset and evaluates on another.
Our experiments follow most previous methods~\cite{Liao_2022_CVPR}, evaluating DG settings on Market1501, MSMT17, CUHK03-NP and DukeMTMC.
In addition, we conduct within-dataset evaluations to assess the discriminative ability of our model, and compare it with several recent DG ReID methods.
As shown in Tab.~\ref{table:DG sota}, our model significantly surpasses other state-of-the-art DG ReID methods.
Notably, under the transfer settings from Market1501 to MSMT17 and MSMT17 to CUHK03-NP, our method outperforms representative methods like OGNorm~\cite{chen2024multi} and PAT~\cite{Ni_2023_ICCV}) by 10.1\% and 11.8\% in mAP, respectively.
At the same time, when trained on Market1501 and tested on DukeMTMC, our method surpasses PAT by 3.9\% and 5.7\% in terms of R1 and mAP, respectively.
When trained on MSMT17, our method outperforms CLIP-ReID by 1.4\% and 2.8\% (test on CUHK03-NP), 0.3.\% and 1.7\% (test on DukeMTMC) in terms of R1 and mAP, respectively.
These results demonstrate the effectiveness of our framework in extracting discriminative features across diverse person ReID scenarios.
\begin{table*}[]
\small
    \caption{Performance comparisons between our method and other methods in both traditional ReID and DG ReID on Market1501, RandPerson, MSMT17, DukeMTMC, and CUHK03-NP.}
    \centering
    \resizebox{0.84\textwidth}{!}{
    \begin{tabular}{l|c|cc|cc|cc|cc}
    \hline
    \multirow{2}{*}{Method}&
    \multirow{2}{*}{Training}&
    \multicolumn{2}{c|}{Market1501}&
    \multicolumn{2}{c|}{MSMT17}&
    \multicolumn{2}{c|}{CUHK03-NP}&
    \multicolumn{2}{c}{DukeMTMC}
    \\
    &       &      mAP         &R1         & mAP         & R1         & mAP         & R1  & mAP         & R1\\
    \hline
    QAConv~\cite{liao2020interpretable}   &\multirow{8}{*}{Market1501}         &--        &--          &7.0         &22.6           &8.6         &9.9  &33.6  &54.4  \\
    TransMatcher~\cite{liao2021transmatcher}      &     & --        &--          &18.4         &47.3           &21.4         &22.2 &--  &--  \\
    QAConv+Gs~\cite{Liao_2022_CVPR}&     &75.5   &91.6   &17.2  &45.9  &18.1  &19.1 &--  &--  \\
    MDA~\cite{Ni_2022_CVPR} &              &--        &--          &11.8         &33.5           &--         &-- &34.4  &56.7  \\
    PAT~\cite{Ni_2023_ICCV} &           &{{81.5}}        &92.4          &18.2         &42.8           &{26.0}        &25.4 &{{48.9}} &{{67.9}}\\
    OGNorm~\cite{chen2024multi} &    &-- &--  &{{19.9}}  &{\underline{49.7}} &{24.9}  &{{26.6}}  &--  &--\\
    CLIP-ReID~\cite{li2023clip}  & &{\underline{90.5}}  &{\underline{95.4}}  &{\underline{23.0}}  &{\underline{48.9}} &{\underline{38.5}} &{\underline{39.9}}  &{\underline{51.7}} &{\underline{69.6}} \\
    % Baseline  & &\cellcolor{gray!40}87.1 &\cellcolor{gray!40}94.1 &10.5 &26.7 &27.2 &29.2 &41.2 &60.6\\
    Ours    &      &{\textbf{91.2}}        &{\textbf{96.0}}         &{\textbf{30.1}}      &{\textbf{58.0}}         &{\textbf{41.4}}    &{\textbf{43.0}} &{\textbf{54.6}}  &{\textbf{71.8}}  \\
    \hline
    QAConv~\cite{liao2020interpretable}   &\multirow{8}{*}{MSMT17}         &43.1        &72.6          &--         &--           &22.6         &25.3 &{{53.4}}  &{\textbf{72.2}}  \\
    TransMatcher~\cite{liao2021transmatcher}    &  &52.0        &{\underline{80.1}}               &--         &--           &22.5         &23.7  &--  &--  \\
    QAConv+Gs~\cite{Liao_2022_CVPR} &     &49.5   &79.1   &50.9  &{{79.2}}  &20.6  &20.9  &--  &--  \\
    MDA~\cite{Ni_2022_CVPR} &           &53.0        &{79.7}          &--         &--           &--         &--  &52.4  &71.7  \\
    PAT~\cite{Ni_2023_ICCV} &           &47.3        &72.2          &{{52.0}}         &75.9           &25.1        &24.2  &--  &--  \\
    OGNorm~\cite{chen2024multi} &  &{\underline{54.5}}  &{\textbf{83.0}}  &-- &--  &{{28.5}}  &{{31.0}}  &--  &--\\
    CLIP-ReID~\cite{li2023clip}  & &{{51.5}}  &{76.2}  &{\underline{75.8}}  &{\underline{89.7}}  &{\underline{38.9}} &{\underline{40.2}}  &{\underline{58.0}}  &{\underline{74.9}} \\
    Ours    &      &{\textbf{55.7}}        &{{80.0}}         &{\textbf{77.0}}      &{\textbf{89.8}}         &{\textbf{41.3}}         &{\textbf{43.0}} &{\textbf{60.3}}  &{\textbf{75.2}}\\
    \hline

    QAConv~\cite{liao2020interpretable}      &\multirow{7}{*}{Multi-source}        &39.5        &68.6          &10.0         &29.9           &19.2         &22.9 &43.4  &64.9  \\
    RaMoE~\cite{dai2021generalizable}    &  &56.5  &82.0  &13.5  &34.1  &{{35.5}} &{{36.6}}  &{\underline{56.9}} &{\underline{73.6}}\\
    $\mathrm{M}^{3}\mathrm{L}$~\cite{zhao2021learning} &           &50.2        &75.9          &14.7         &36.9           &32.1       &33.1 &51.1  &69.2  \\
    CINorm~\cite{chen2023cluster} & &{{57.8}}       &{{82.3}} &21.1  &{{49.7}}  &31.1  &30.3  &52.4  &71.3\\
    PAT~\cite{Ni_2023_ICCV}  & & 51.7 &75.2 &{{21.6}} &45.6 &31.5 &31.1 &{{56.5}} &{{71.8}}\\
    OGNorm~\cite{chen2024multi} &  &{\textbf{65.2}}  &{\textbf{87.1}}  &{\underline{25.9}}  &{\underline{57.7}}  &{\underline{40.3}}  &{\underline{44.0}}  &--  &--\\
    % Baseline  & & --& --& --& --& --& --& --& --\\
    Ours    &      &{\underline{61.6}}        &{\underline{83.1}}         &{\textbf{29.7}}      &{\textbf{58.7}}         &{\textbf{42.0}}         &{\textbf{42.9}} &{\textbf{61.6}}  &{\textbf{76.3}}\\
    \hline
    QAConv+Gs~\cite{Liao_2022_CVPR}&\multirow{4}{*}{RandPerson}     &46.7   &76.7   &15.5  &45.1  &18.4  &16.1 &--  &--  \\
    PAT~\cite{Ni_2023_ICCV} &           &{{46.9}}        &{{73.7}}          &{\underline{19.4}}         &{{45.5}}           &{\underline{20.2}}        &{{20.1}} &--  &--  \\
    OGNorm~\cite{chen2024multi} &  &{\underline{50.2}}  &{\underline{78.9}}  &{{19.3}}  &{\underline{49.8}}  &{{19.1}}  &{\underline{21.5}}  &--  &--\\
    % Baseline  & & --& --& --& --& --& --& --& --\\
    Ours    &      &{\textbf{59.2}}        &{\textbf{81.0}}         &{\textbf{27.4}}      &{\textbf{56.4}}         &{\textbf{34.8}}         &{\textbf{36.6}} &{\textbf{54.1}}  &{\textbf{73.8}}\\
    \hline
    \end{tabular}}
    \label{table:DG sota}
\end{table*}

\textbf{Multi-source DG ReID.}
To further validate the generalization ability of our model, we also present results in the multi-source setting.
As shown in Tab.~\ref{table:DG sota}, our method outperforms other methods on most datasets.
Specifically, our model's mAP is 3.8\% higher than the current best model (OGNorm) in the \textbf{M}+\textbf{D}+\textbf{C}$\rightarrow$ \textbf{MS} setting and reaches 42.0\% in the \textbf{MS}+\textbf{M}+\textbf{D}$\rightarrow$ \textbf{C} setting.
When training on the \textbf{M}+\textbf{MS}+\textbf{C}, our model also achieves excellent results which outperform PAT~\cite{Ni_2023_ICCV} by a large margin with 5.1\% in mAP.

\textbf{Synthetic-to-Real DG ReID.}
In addition, we evaluate the model's generalization ability by training on synthetic data and testing on real data.
As shown in Tab.~\ref{table:DG sota}, our method significantly outperforms all state-of-the-art methods.
Specifically, when trained on RandPerson and evaluated on CUHK03, our model achieves 15.7\% and 15.1\% improvements in terms of mAP and R1, respectively, compared with the current best method, OGNorm.
Notably, despite trained solely on synthetic data, our method delivers competitive performances even when compared with methods trained directly on real-world datasets (\emph{e.g.}, OGNorm).
These results highlight the superior generalization ability of our method to unseen domains.

As demonstrated in previous works~\cite{chen2023cluster,chen2024multi}, performances degrade significantly when tested on unseen domains.
To address this issue, most DG ReID methods introduce meta-learning and normalization techniques to learn more generalized features.
However, these techniques inevitably neglect fine-grained cues, as evidenced by the poor performances of these methods in single-domain settings.
In contrast, our method decouples visual representations into multiple attribute prompts.
These prompts are then combined to obtain attribute-aware features, which yield more robust representations.

\textbf{Text-based Person Retrieval.}
Our framework can be easily extended to text-based person retrieval.
Tab.~\ref{tab:text-reid} shows compared results with state-of-the-art methods on CUHK-PEDES~\cite{li2017person} and ICFG-PEDES~\cite{ding2021semantically}.
As observed, our method is also effective for text-based person retrieval, highlighting the effectiveness of attribute learning.
These results clearly show the generalization ability of our method.
%-------------------------------------------------
\begin{table}[hbtp]
\centering
\caption{Quantitative comparison of state-of-the-art methods on CUHK-PEDES and ICFG-PEDES.}
\resizebox{0.46\textwidth}{!}{
\begin{tabular}{c|cccc|cccc}
        \hline
        \multirow{2}{*}{Model} &\multicolumn{4}{c|}{CUHK-PEDES} &\multicolumn{4}{c}{ICFG-PEDES}\\
         & mAP & R1 & R5 &R10 & mAP & R1 & R5 &R10 \\
        \hline
         LGUR~\cite{LGUR}  &- &65.2 &83.1 &89.0 &- &57.4 &75.0 &81.5 \\
         IVT~\cite{IVT}  &60.7 &65.6  &83.1  &89.2   &-  &56.0  &73.6  &80.2\\
         SAF~\cite{SAF} &58.6 &64.1  &82.6  &88.4  &32.8 &54.9  &72.1  &79.1\\
         BLIP~\cite{blip}  &58.0 &65.6  &82.8  &88.65 &38.8 &56.1  &75.7  &82.8\\
         EAIBC~\cite{EAIBC}   &- &65.0 &83.3 &88.4 &- &58.9  &76.0  &81.7\\
         FedSH~\cite{FedSH}  &-  &60.9   &80.8  &87.6 &- &55.0  &72.8  &79.5\\
         TBPS-CLIP~\cite{cao2024empirical}   &64.9 &72.6 &88.1 &92.7  &39.5  &64.5  &80.0  &85.4\\

        \hline
         Ours  & \textbf{65.2}      &\textbf{73.0}   &\textbf{88.6}  &\textbf{93.1}  &\textbf{40.0}  &\textbf{64.9}  &\textbf{80.6}  &\textbf{85.9}\\
        \hline
\end{tabular}}
\label{tab:text-reid}
\end{table}
\subsection{Ablation Studies}
We employ the standard identity loss and triplet loss to train the baseline model which bases on the ViT-B/16 from CLIP with patch overlapping and camera information.

\textbf{Ablation Study of Main Components.}
As shown in Tab.~\ref{table:ablation study}, to evaluate the contribution of each component, we conduct ablation studies on MSMT17 under both single-domain setting and DG setting.
Firstly, in the experiment (b), we utilize SUS to generate the visual identity prototype, and remove the textual representations.
As the results demonstrated, FSTS preserves the image-level perceptual ability inherited from VLM, while effectively adapting to ReID-specific discriminative learning.
Secondly, we further introduce the SAD to obtain textual representations.
Specifically, we calculate the cosine similarity between the attribute prompts and visual representations, and use it to aggregate attribute prompts into attribute-aware features.
As shown in experiment (c), this design improves performances in both single-domain and DG settings.
This indicates that the attribute-aware features are both discriminative and robust across domains.
Finally, as shown in experiment (d), when combining FSTS and APG, the performances are further enhanced in both settings.
This indicates that the model obtains both generalizable and discriminative representations, which capture the relationships between attribute prompts and visual representations.
\begin{table}[hbtp]
\centering
\caption{Performance analysis of each component.}
\resizebox{0.46\textwidth}{!}{
\begin{tabular}{c|cccc|cccc}
\hline
&\multirow{2}{*}{Base}  &\multirow{2}{*}{FSTS} &\multicolumn{2}{c|}{APG} &\multicolumn{2}{c}{\textbf{MS}} &\multicolumn{2}{c}{\textbf{MS}$\rightarrow$ \textbf{M}} \\
&  &  &SAD  &PCM  &mAP &R1 &mAP &R1 \\
\hline
(a)&\checkmark &$\times$ &$\times$ &$\times$      &66.4  &84.4 &16.2  &32.5\\
(b)&\checkmark &\checkmark &$\times$   &$\times$   &75.0  &88.8 &42.2  &67.6\\
(c)&\checkmark &\checkmark &\checkmark &$\times$    &76.8  &89.5 &46.5 &72.6\\
(d)&\checkmark &\checkmark &\checkmark &\checkmark  &\textbf{77.1}  &\textbf{90.1} &\textbf{55.7}  &\textbf{80.0} \\
\hline
\end{tabular}}
\label{table:ablation study}
\end{table}

\textbf{Effect of Learned Textual Attributes.}
To validate that SAD effectively learns meaningful attributes, we conduct ablation experiments in Tab~\ref{tab:prototype prompt textual}.
First, to examine whether textual information contributes to feature representations, we replace SAD with randomly initialized vectors.
The model's performances drop significantly, indicating that the absence of textual semantics leads to overfitting and redundant representations.
Second, we manually define 113 attributes covering age, gender, hairstyle, upper and lower clothing, and backpacks.
When the learnable attributes are replaced with manually designed ones, the performance is substantially reduced.
This suggests that incorporating textual semantics provides beneficial guidance for object representations.
However, as handcrafted attributes may not fully capture the diversity of attributes, the overall performances remain slightly inferior.
In contrast, learning attribute prompts adaptively through SAD yields the best results.
These results validate that SAD effectively aligns visual features with diverse object attributes, ultimately resulting in more robust representations.
\begin{table}[hbtp]
\centering
\caption{Analysis of learned attribute prompts.}
\resizebox{0.46\textwidth}{!}{
\begin{tabular}{c|cc|cc}
\hline
\multirow{2}{*}{} & \multicolumn{2}{c|}{MSMT17} & \multicolumn{2}{c}{DukeMTMC}  \\
& mAP          & R1          &mAP            &R1  \\
\hline
Directly Learning   & 74.9      &89.1   &82.4            &91.0\\
Manually Designed  & 76.0      &89.6   &83.6    &91.5\\
Learned Attribute Prompts (Ours)  & \textbf{77.1}      &\textbf{90.1}   &\textbf{84.1}    &\textbf{91.9}\\
\hline
\end{tabular}
}
\label{tab:prototype prompt textual}
\end{table}
%-----------------------------------------------
\begin{table}[hbtp]
\centering
\caption{Performance analysis with varying numbers of attribute prompts in SAD.}
\resizebox{0.30\textwidth}{!}{
\begin{tabular}{c|cc|cc}
\hline
\multirow{2}{*}{No.} & \multicolumn{2}{c|}{MSMT17} & \multicolumn{2}{c}{DukeMTMC}  \\
& mAP          & R1          &mAP            &R1  \\
\hline
$8$  & 71.8      &87.6   &79.8     &89.7\\
$16$  & 74.0      &88.4   &82.1    &90.5\\
$32$  & 75.4      &89.0   &83.4    &91.0\\
$64$  & 76.2      &89.6   &84.0    &91.8\\
$128$  & 76.9      &89.8   &84.3    &91.9\\
$256$  & 77.1      &90.1   &84.1    &91.9\\
$512$  & 77.0      &89.9   &84.2    &91.7\\
\hline
\end{tabular}}
\label{tab:learnable attributes}
\end{table}

\textbf{Effect of the Number of Attribute Prompts in SAD.}
As shown in Tab.~\ref{tab:learnable attributes}, we present the performances with varying numbers of attribute prompts in SAD.
The training is conducted on the MSMT17 and DukeMTMC datasets.
Specifically, we vary the number of attribute prompts from $8$ to $512$.
As illustrated in Tab.~\ref{tab:learnable attributes}, the performances on MSMT17 and DukeMTMC consistently improve with an increase of the number of learnable attributes.
However, as the number of attributes increases, the performances approach saturation. To ensure a fair comparison and parameter consistency across all datasets, we fix the number of attributes to 256.

\textbf{Effect of the Number of Learnable Tokens for Each Attribute.}
As shown in Tab.~\ref{tab:attribute prompt tokens}, we report the results with varying numbers of learnable tokens for each attribute.
The experiments are conducted on both MSMT17 and DukeMTMC.
As observed, the results across both datasets remain relatively stable when increasing the number of learnable tokens.
Therefore, we adopt 4 tokens in our model as a trade-off between performance and efficiency.
\begin{table}[hbtp]
\centering
\caption{Effect of learnable tokens for each attribute.}
\resizebox{0.32\textwidth}{!}{
\begin{tabular}{c|cc|cc}
\hline
\multirow{2}{*}{} & \multicolumn{2}{c|}{MSMT17} & \multicolumn{2}{c}{DukeMTMC}  \\
& mAP          & R1          &mAP            &R1  \\
\hline
1  & 76.7      &89.5   &84.2            &91.1\\
2  & 76.6      &89.9   &84.2    &91.9\\
4  & \textbf{77.1}      &\textbf{90.1}   &\textbf{84.1}    &\textbf{91.9}\\
8  & 76.9      &90.0   &84.2    &91.9\\
16 & 77.0      &90.1   &83.9            &91.8\\
\hline
\end{tabular}
}
\label{tab:attribute prompt tokens}
\end{table}
%-------------------------------------
\begin{table}[hbtp]
\centering
\caption{Effect of the orthogonality loss.}
\resizebox{0.32\textwidth}{!}{
\begin{tabular}{c|cc|cc}
\hline
\multirow{2}{*}{Setting} & \multicolumn{2}{c|}{MSMT17} & \multicolumn{2}{c}{DukeMTMC}  \\
& mAP          & R1          &mAP            &R1  \\
\hline
w/o $\mathcal{L}_{\text{orth}}$   & 76.1      &89.0   &83.3            &90.5\\
w $\mathcal{L}_{\text{orth}}$ & \textbf{77.1}      &\textbf{90.1}   &\textbf{84.1}    &\textbf{91.9}\\
\hline
\end{tabular}
}
\label{tab:orthogonality}
\end{table}

\textbf{Effect of the Orthogonality Loss.}
To encourage the framework to capture more diverse and complementary features, we introduce an orthogonality loss to the attribute prompts.
As shown in Tab.~\ref{tab:orthogonality}, with the orthogonality loss, our framework consistently improves performance on both MSMT17 and DukeMTMC.
It enhances the diversity of learned attributes and thereby boosts the overall model performance.

\textbf{Computational Cost Analysis of Our APC.}
Tab.\ref{tab:comp_cost} shows the comparison of computation and performance between the baseline and our APC on MSMT17.
During inference, SAD is invoked once to generate the shared attribute prompts, and the feature extraction relies solely on SUS.
Although multiple interactions occur between the visual features and attribute prompts, the additional overhead is minor.
As shown in Tab.~\ref{tab:comp_cost}, our method achieves 579.8 images/s compared with 588.9 images/s for the baseline.
It only has a slight increase in parameters and memory usage.
Importantly, our method yields a significant performance gain of +10.7 mAP on MSMT17.
\begin{table}[hbtp]
\centering
\caption{Comparison of computation and performance between the baseline and our APC on MSMT17.}
\begin{tabular}{c|cccc}
\hline
Model & Speed & Param. & Memory &mAP \\
\hline
Baseline  & 588.9 images/s     &86.2 M   &5,385 MB  &66.4\\
Ours   & 579.8 images/s     &94.4 M   &5,431 MB  &77.1\\
\hline
\end{tabular}
\label{tab:comp_cost}
\end{table}
%------------------------------------------------------------
\begin{figure}[htbp]
\centering
\includegraphics[width=0.9\linewidth]{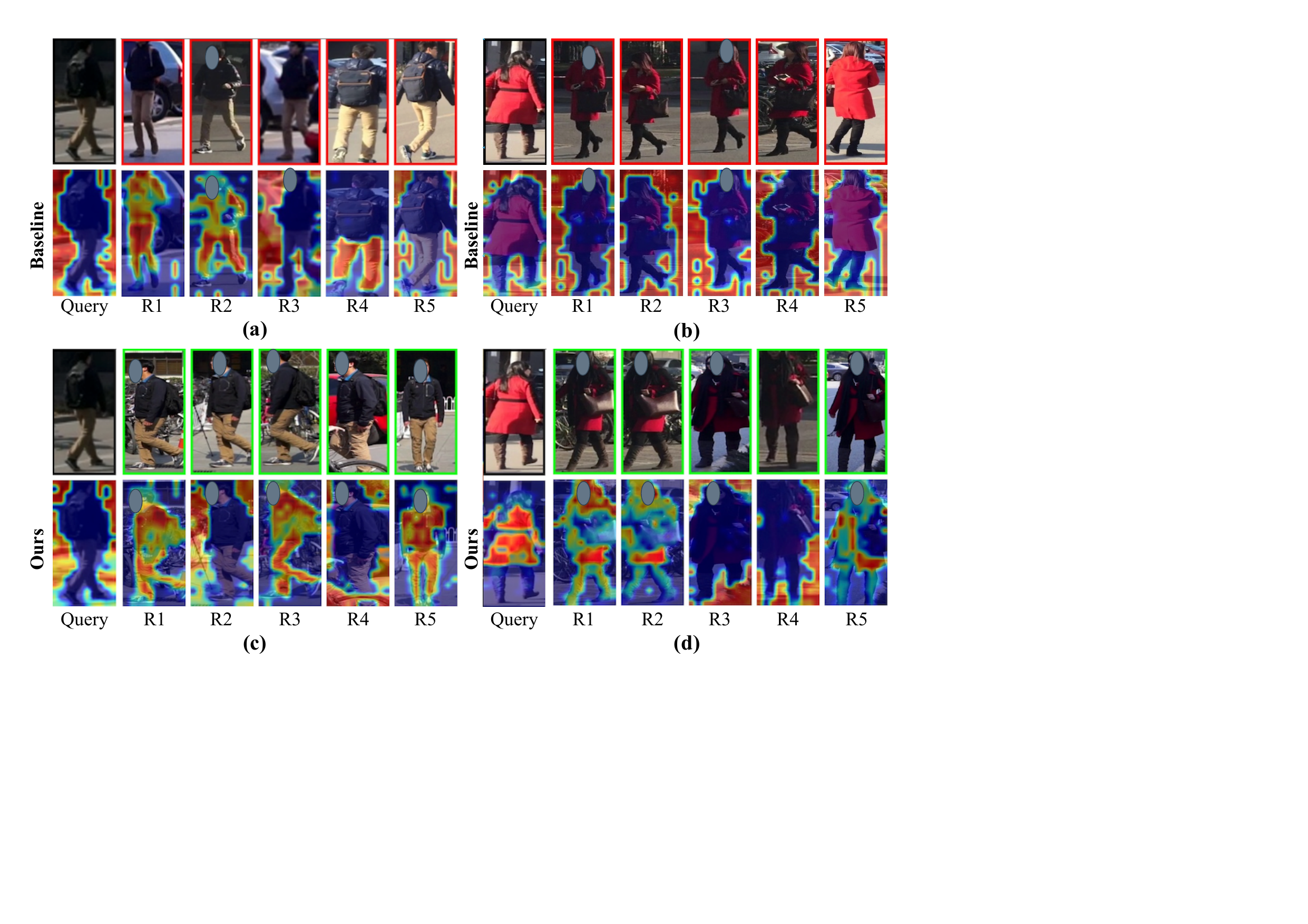}
\caption{Heat maps and retrieval rank lists with the baseline and our method. Deeper red colors signify higher weights.}
\label{fig:3}
\end{figure}
\subsection{Qualitative Results}
To better understand our proposed framework, we present comprehensive qualitative results in this section.

\textbf{Top-5 Retrieval and Heatmap Results.}
Fig.~\ref{fig:3} shows two examples of top-5 retrieval results along with their heat maps generated by EigenCAM~\cite{EigenCAM}.
Specifically, Fig.~\ref{fig:3} (a) and (b) display the rank list and heatmap results with the baseline, while Fig.~\ref{fig:3} (c) and (d) show the rank list and heatmap results with our methods for the same queries.
It can be observed that the baseline struggles to distinguish samples that have similar appearances but subtle differences in details.
The main reason is that visual-only models tend to overfit to prominent appearance cues, while neglecting the unique attributes of individuals.
For example, as shown in Fig.~\ref{fig:3} (a) R1 and Fig.~\ref{fig:3} (a) Rank-2 (R2), although the model highlights the correct regions of the pedestrians, the retrieval results are still incorrect.
This is because the model only focuses on prominent cues such as the black coat and brown pants, while neglecting other detailed attributes like gender and bag color.
As shown in Fig.~\ref{fig:3} (c), by incorporating SAD, APC focuses more on detailed visual cues and avoids overfitting to the prominent appearance information.
The same phenomenon can be observed in Fig.~\ref{fig:3} (b) and Fig.~\ref{fig:3} (d), where our method successfully recognizes detailed attribute information, such as hair color and clothing style.
\begin{figure}[htbp]
\centering
\includegraphics[width=0.9\linewidth]{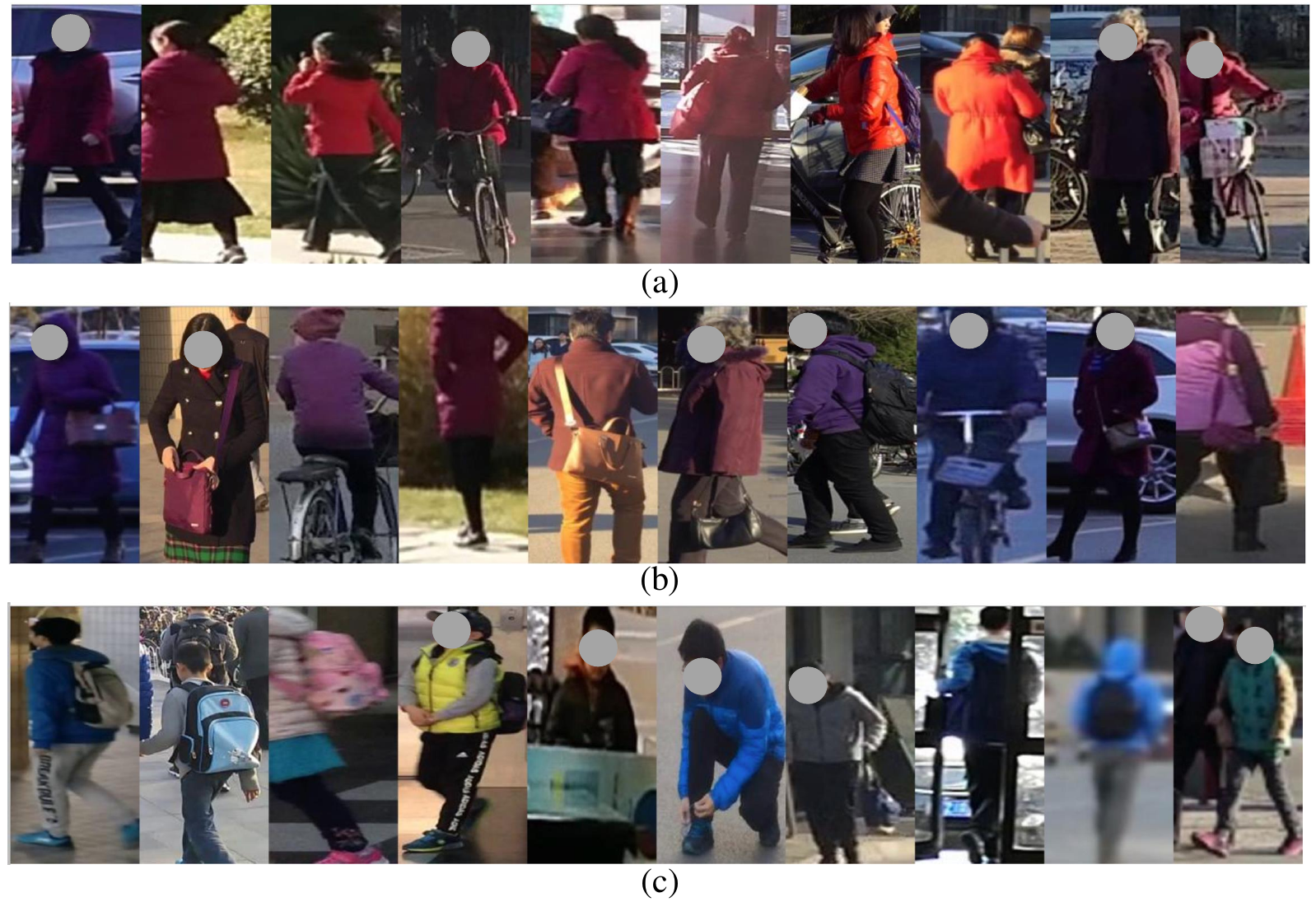}
\caption{Retrieval results using different attribute prompts on MSMT17. Each row shows images closest to selected attribute.}
\label{fig:4}
\end{figure}

\begin{figure}[htbp]
\centering
\includegraphics[width=0.9\linewidth]{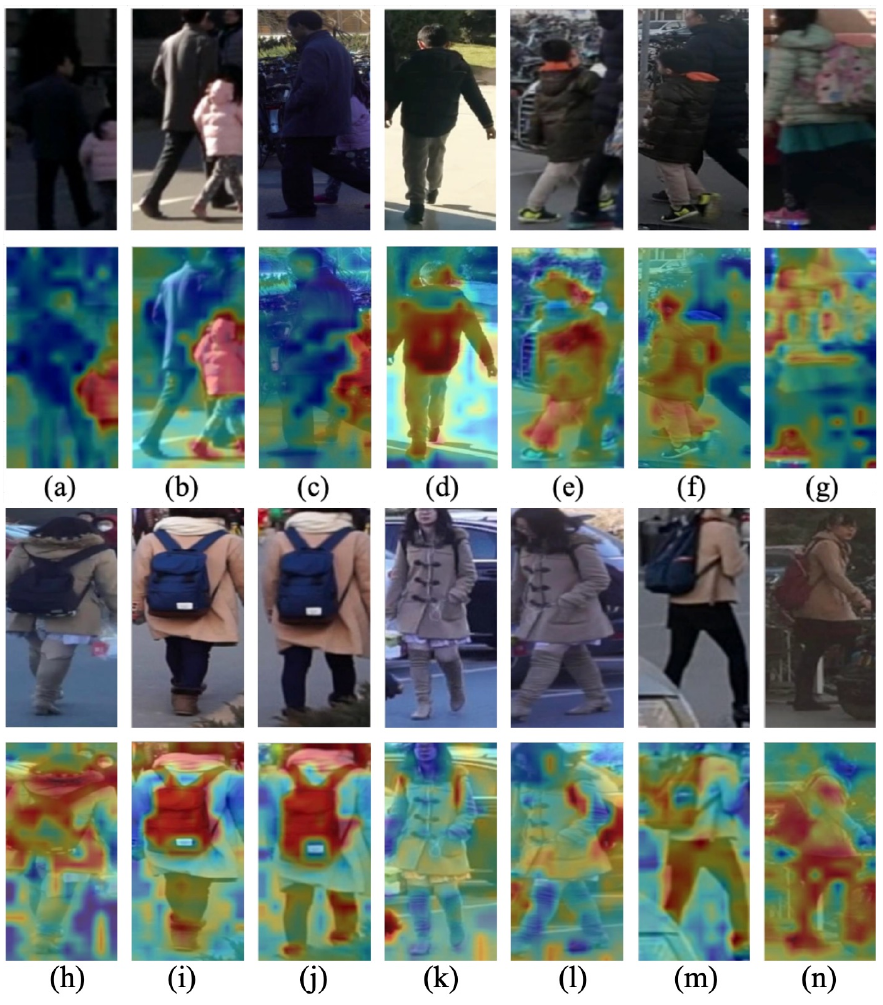}
\caption{Visualizations of similarities between the patch tokens and attribute prompt. Each row shares an attribute prompt.}
\label{fig:5}
\end{figure}

\begin{figure}[htbp]
\centering
\includegraphics[width=0.44\linewidth]{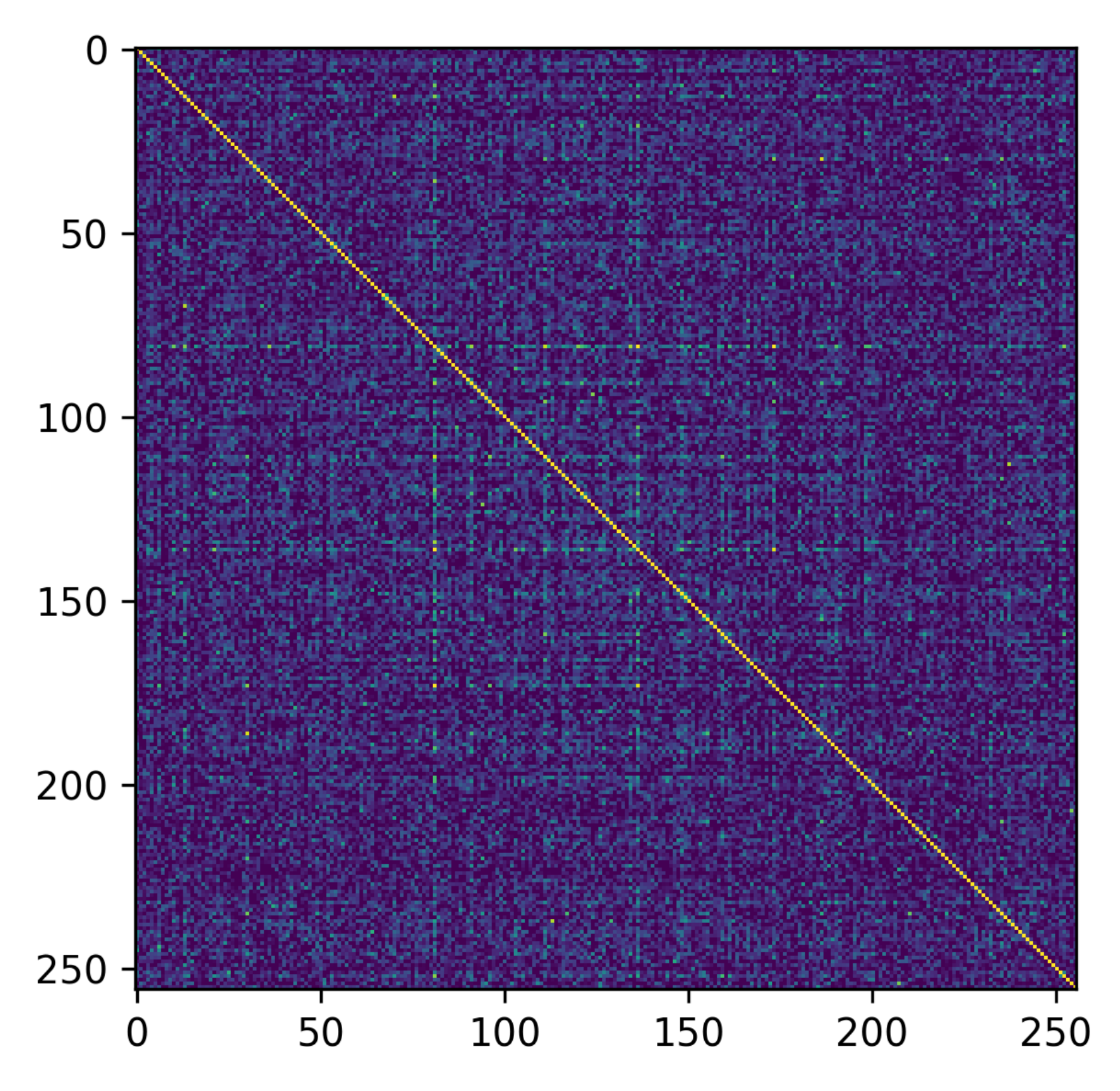}
\caption{Row-normalized Gram matrix of learnable attributes.}
\label{fig:gram}
\end{figure}

\textbf{Interpretation of Learned Attribute Prompts.}
We further explore the interpretability of learned attribute prompts.
More specifically, we compute the cosine similarity between each attribute prompt $g_j$ and the projected visual representations $r$, and retrieve the top-ranked images to construct a semantic retrieval list.
As illustrated in Fig.~\ref{fig:4}, each row corresponds to the retrievals associated with a specific attribute.
It can be observed that each attribute captures a distinct pedestrian-related semantic.
For instance, the attribute prompts in Fig.~\ref{fig:4} (a), Fig.~\ref{fig:4} (b) and Fig.~\ref{fig:4} (c) can be interpreted as representing ``female with a red coat'', ``female with a purple coat'' and ``teenager'', respectively.
In this work, we employ the orthogonality loss to encourage SAD to enlarge the similarity among learnable attribute prompts in the feature space.
It makes the model focus more on salient attributes.
However, it cannot completely remove the influence of related attributes.
As shown in Fig.~\ref{fig:4}, attributes such as \emph{red coat} and \emph{purple coat} tend to co-occur with female.
Thus, modeling these attributes captures features related to female.
Fig.~\ref{fig:gram} shows the learned relationships of different attributes, which hold non-zero off-diagonal entries.
To further assess whether the attribute prompts accurately localize semantic concepts, we visualize their spatial correspondence with image regions.
As shown in Fig.~\ref{fig:5}, each row presents pixel-wise heat maps of the cosine similarity between the attribute prompt and different images.
The highlighted regions consistently correspond to the same semantic concept across images.
For example, the visualized semantics in Fig.~\ref{fig:5} include ``child'' and ``blue backpack'', respectively.
Notably, as shown in Fig.~\ref{fig:5} (j) and Fig.~\ref{fig:5} (k), the attribute prompt accurately attends to backpack-related regions even across different identities, including fine-grained details such as shoulder straps.
This further confirms the semantic consistency and localization precision of the learned attribute prompts.
In summary, the visualization results demonstrate that the learned attribute prompts encode clear and consistent object-related semantics.
This enhances the robustness and generalization of the feature representations, while helping to mitigate overfitting.

\begin{figure}[htbp]
\centering
\includegraphics[width=0.9\linewidth]{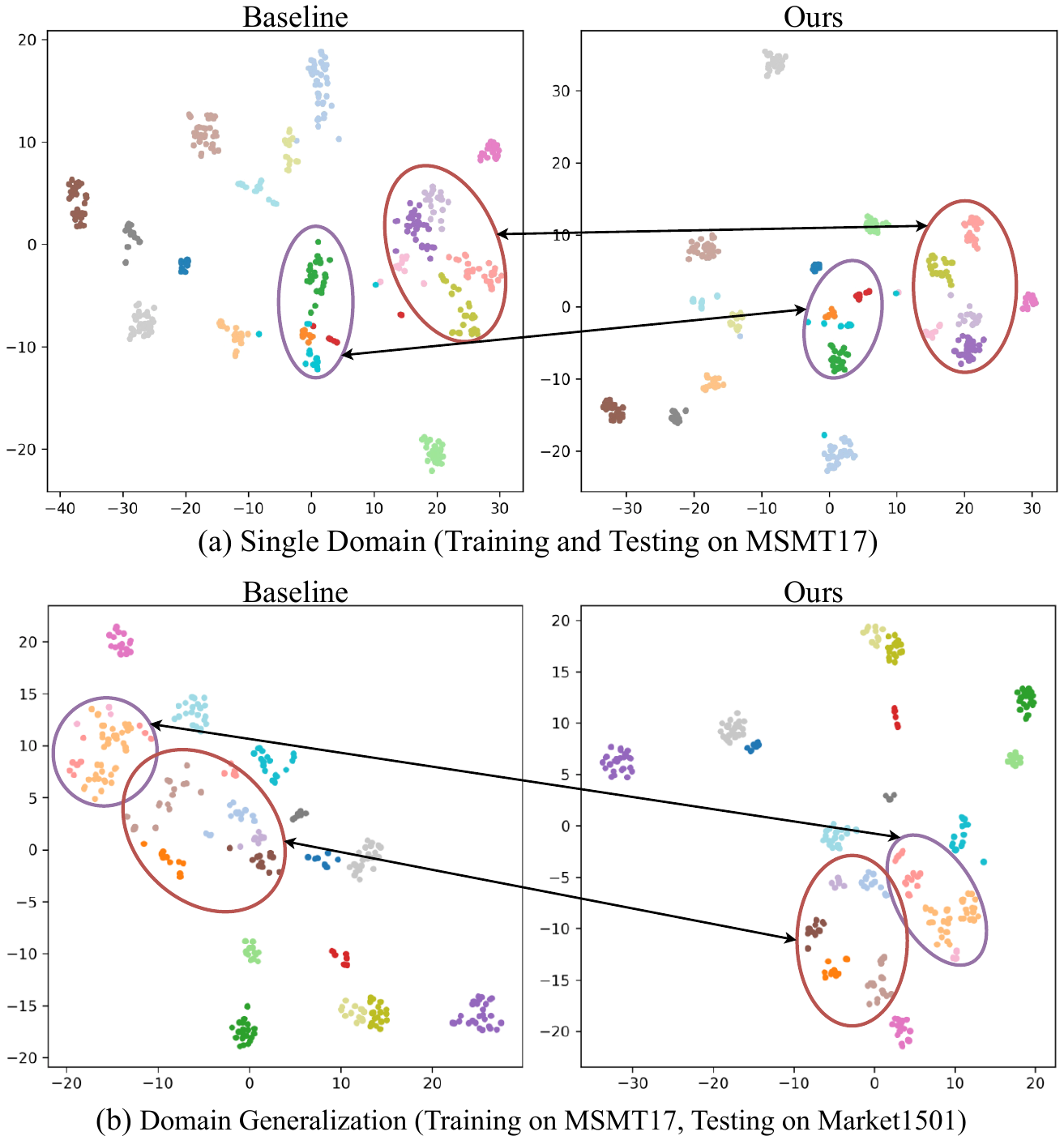}
\caption{Visualization of feature distributions with the baseline and our method under different settings. Each point represents a sample, and each color indicates an identity.}
\label{fig:6}
\end{figure}

\textbf{Visualization of Feature Distributions.}
In Fig.~\ref{fig:6}, we visualize the feature distributions using t-SNE~\cite{tsne} under both single-domain and DG settings, .
In the single-domain scenario (Fig.~\ref{fig:6} (a)), compared to the baseline, our method significantly enhances the intra-identity compactness and inter-identity separability.
In the DG setting (Fig.~\ref{fig:6} (b)), the baseline features exhibit a severe domain shift with overlapping clusters, while our method delivers more discriminative and well-separated embeddings despite the domain gap.
These visualizations clearly demonstrate the superiority of our method in learning domain-robust and identity-consistent representations.

\textbf{Visualization of Attribute Orthogonality.}
To verify whether learnable attributes are distributed orthogonally, we visualize the row-normalized Gram matrix of attribute tokens.
As shown in Fig.~\ref{fig:gram}, it clearly shows that learned attributes are approximately orthogonal.
This orthogonality further enhance the diversity of the SAD representations.
Note that there are non-zero off-diagonal entries.
It indicates that specific relations of different attributes still exist, such as long hair and female.
\section{Conclusion}
\label{sec:conclusion}
In this paper, we propose a novel Attribute Prompt Composition (APC) framework that integrates discriminative visual representations with generalized textual semantic information for image-based object ReID.
More specifically, we first introduce an Attribute Prompt Generator (APG) to represent an object as a composition of multiple attributes.
The APG consists of two key components: Semantic Attribute Dictionary (SAD) and Prompt Composition Module (PCM).
SAD is an over-complete attribute dictionary to provide rich semantic descriptions, while PCM adaptively composes relevant attributes from SAD to generate discriminative attribute-aware features.
Furthermore, we incorporate a Fast-Slow Training Strategy (FSTS) to supervise the learning of our framework.
It captures ReID-specific information with the visual perception ability inherited from the pre-trained VLM.
Extensive experiments on five widely used ReID benchmarks demonstrate that our method significantly outperforms other methods in both traditional ReID and DG ReID tasks.
In the future, we will explore more effective attribute selection methods to improve the feature representation and reduce the computation.
%-----------------------------reference---------------------------------
% \newpage
\bibliographystyle{IEEEtran}
\bibliography{IEEEabrv,main}
%-----------------------------Authors---------------------------------
\begin{IEEEbiography}[{\includegraphics[width=1in,height=1.24in,clip,keepaspectratio]{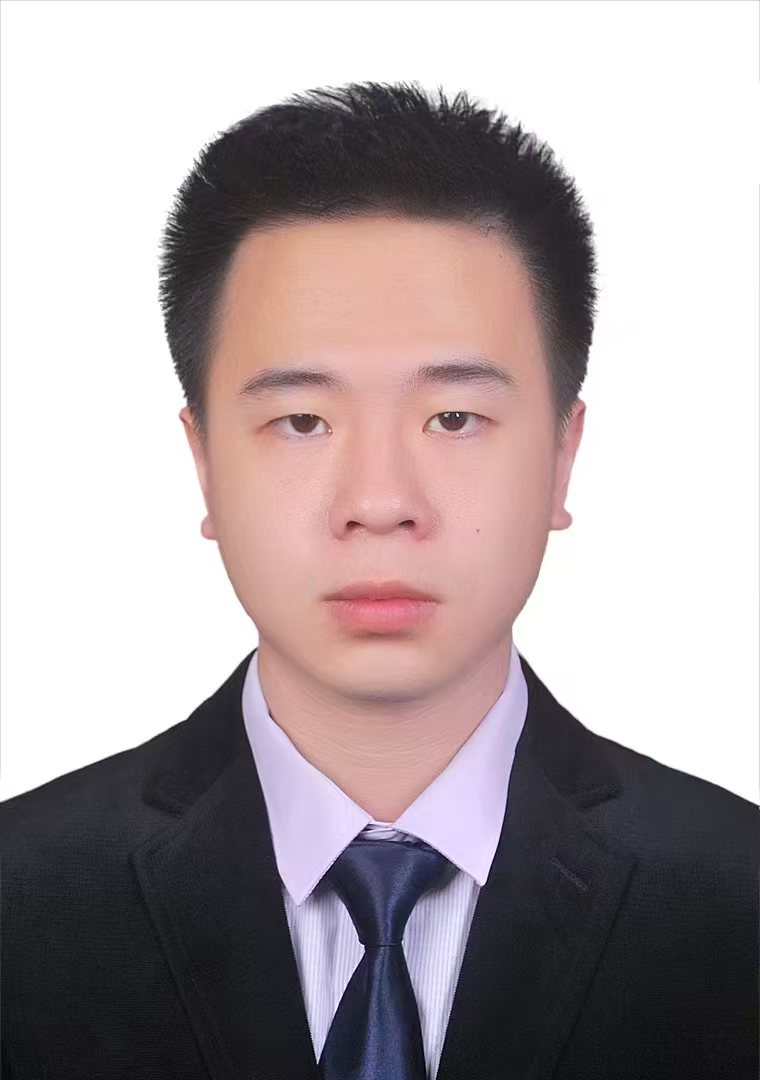}}]{Yingquan Wang} received the B.E. degree in mathematics and applied mathematics from Anhui University, Hefei, China, in 2017, and the M.E. degree in pattern recognition and intelligent systems from Jiangsu University, Zhenjiang, China, in 2022. He is currently pursuing the Ph.D. degree in signal and information processing at Dalian University of Technology (DUT), Dalian, China. His research interests include deep learning and person re-identification.
\end{IEEEbiography}
\vspace{-6mm}
\begin{IEEEbiography}[{\includegraphics[width=1in,height=1.24in,clip,keepaspectratio]{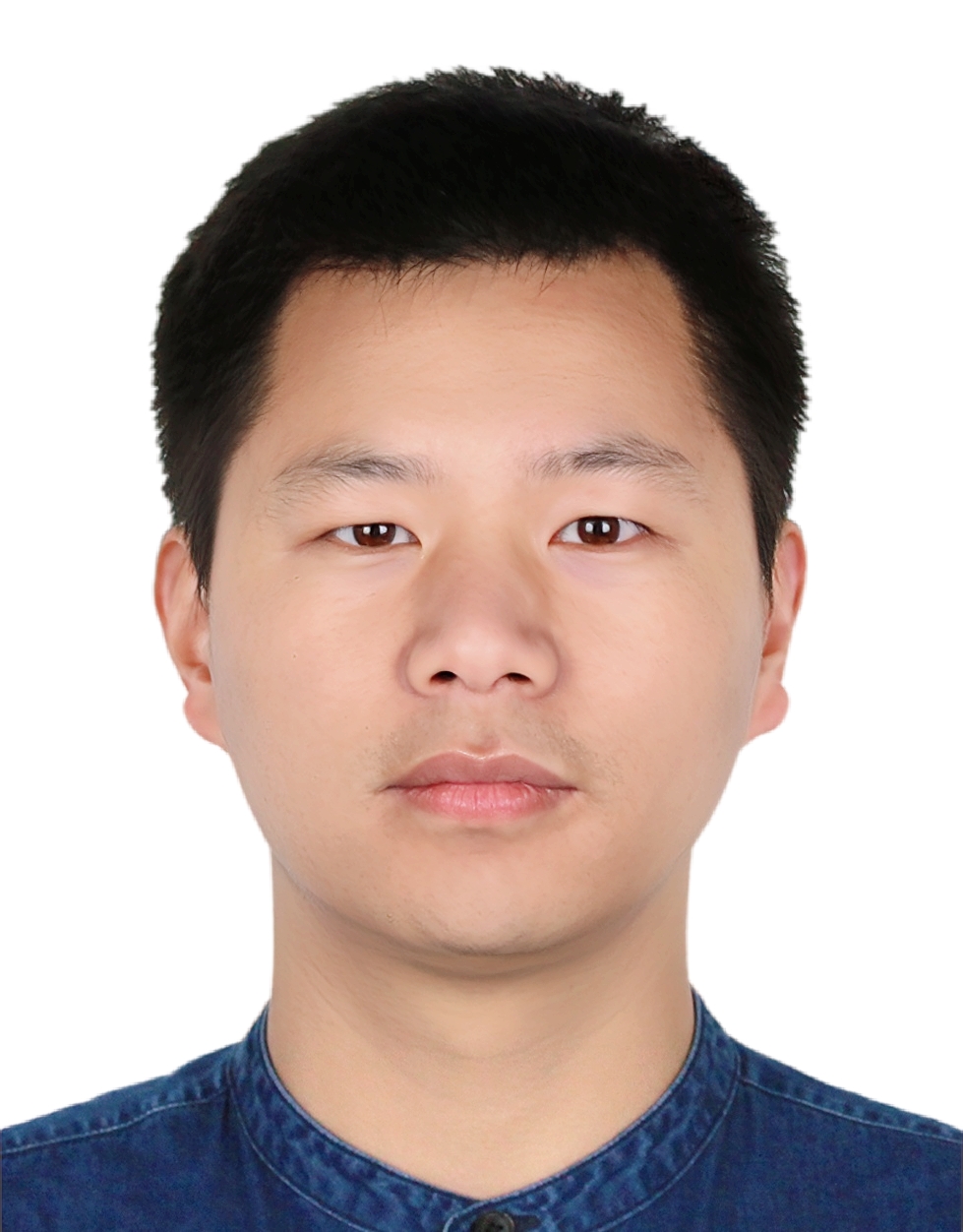}}]{Pingping Zhang} received the B.E. degree in mathematics and applied mathematics from Henan Normal University (HNU), Xinxiang, China, in 2012, and the Ph.D. degree in signal and information processing from the Dalian University of Technology (DUT), Dalian, China, in 2020. He is currently an Associate Professor with the School of Future Technology/School of Artificial Intelligence, DUT. He has authored over 70 top-tier journal/conference papers. His research interests include deep learning, saliency detection, person re-identification, and semantic segmentation.
\end{IEEEbiography}
\vspace{-6mm}
\begin{IEEEbiography}[{\includegraphics[width=1in,height=1.24in,clip,keepaspectratio]{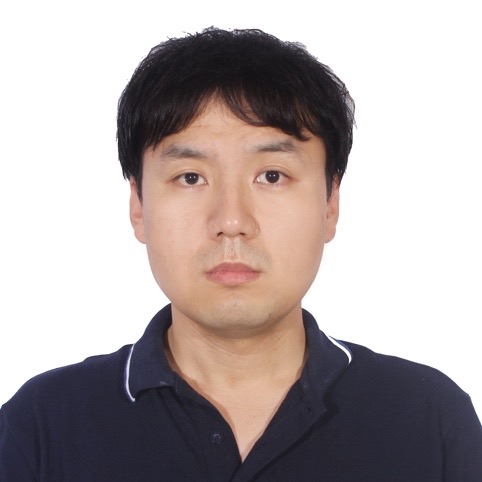}}]{Chong Sun} received the B.E. degree in electronic information engineering and the Ph.D degree in signal and information processing from the Dalian University of Technology (DUT) in 2012 and 2018, respectively. He is currently a senior researcher in WeChat, Tencent. His research interests include multi-modal models, visual object tracking, object detection and person re-identification.
\end{IEEEbiography}
\vspace{-6mm}
\begin{IEEEbiography}[{\includegraphics[width=1in,height=1.24in,clip]{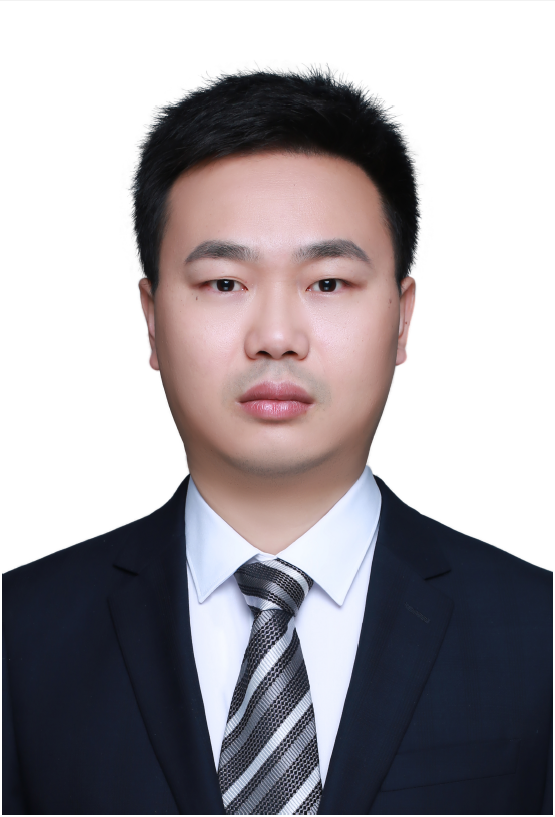}}]{Dong Wang} received the B.E. degree in electronic information engineering and the Ph.D. degree in signal and information processing from Dalian University of Technology (DUT), Dalian, China, in 2008 and 2013, respectively. He is currently a full professor with the School of Information and Communication Engineering, DUT. His current research interests mainly focus on object detection and tracking, and embodied intelligence.
\end{IEEEbiography}
\vspace{-6mm}
\begin{IEEEbiography}[{\includegraphics[width=1in,height=1.24in,clip,keepaspectratio]{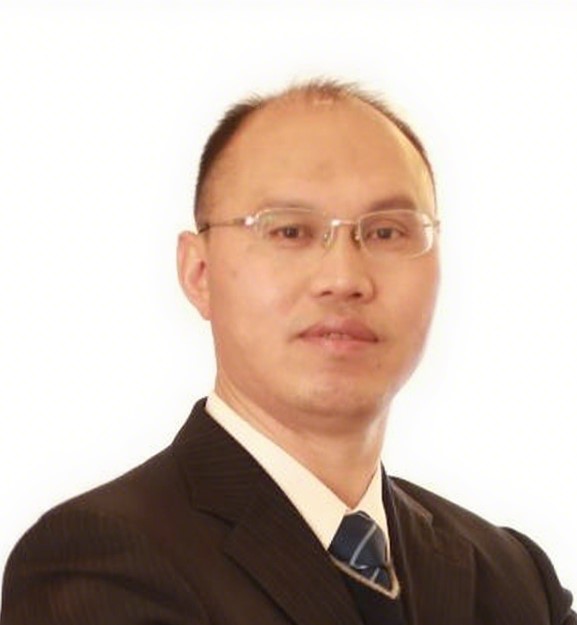}}]{Huchuan Lu} (IEEE Fellow) received the M.S. degree in signal and information processing, Ph.D degree in system engineering, Dalian
University of Technology (DUT), China, in 1998 and 2008 respectively. He has been a faculty since 1998 and a professor since 2012 in the School of Information and Communication Engineering of DUT. His research interests are in the areas of computer vision and pattern recognition. In recent years, he focus on visual tracking, saliency detection and semantic segmentation. Now, he serves as an associate editor of the IEEE Transactions On Cybernetics.
\end{IEEEbiography}
\vspace{-6mm}
% to do
\end{document}